%% file: acl_latex.tex
\pdfoutput=1

\documentclass[11pt]{article}

\usepackage[preprint]{acl}

\usepackage{times}
\usepackage{latexsym}

\usepackage[T1]{fontenc}

\usepackage[utf8]{inputenc}

\usepackage{microtype}

\usepackage{inconsolata}

\usepackage{graphicx}

\usepackage{algorithm}
\usepackage{algorithmic}
\usepackage{amsmath} 
\usepackage{booktabs}
\usepackage{xcolor}
\usepackage[utf8]{inputenc}
\usepackage{tcolorbox}
\usepackage{multirow}

\title{ResoFilter: Fine-grained Synthetic Data Filtering for Large Language Models through Data-Parameter Resonance Analysis}

\author{Zeao Tu$^{1}$\footnotemark[1], Xiangdi Meng$^{1}$\footnotemark[1] \\
\textbf{Yu He}$^{1,2}$\footnotemark[2], 
\textbf{Zihan Yao}$^{1}$, \textbf{Tianyu Qi}$^{1}$, \textbf{Jun Liu}$^{2}$, \textbf{Ming Li}$^{1}$ \\
$^1$TAL Education Group, Beijing, China \\
$^2$Xi’an Jiaotong University, Xi’an, Shaanxi, China\\
\texttt{\{tuzeao, mengxiangdi, heyu26, yaozihan1, qitianyu3, liming9\}@tal.com}\\ 
\texttt{liukeen@mail.xjtu.edu.cn}\\
}

\usepackage{hyperref}

\begin{document}
\maketitle

\input{latex/mxd/abs}
\renewcommand{\thefootnote}{\fnsymbol{footnote}}
\footnotetext[1]{Equal contribution.}
\footnotetext[2]{Corresponding author.}
\input{latex/mxd/intro}


\section{Related Work}
\paragraph{Instruction data selection method}
Using high-quality data for instruction tuning of large language models can significantly enhance their performance. The existing work on selecting high-quality data can be divided into two main parts:
(1).Utilizing human expertise or more powerful generative models to assist in the selection process. For example, Lima~\cite{zhou2024lima} manually curated a dataset of 1,000 instruction samples for instruction tuning and achieved outstanding performance on the evaluation set. AlpaGasus~\cite{chen2023alpagasus}, InsTag~\cite{lu2023instag}, and DEITA~\cite{liu2023makes} leverage the superior performance of ChatGPT to replace manual effort in filtering and selecting data.
(2).Model-guided data selection involves filtering data by observing changes in the model or by introducing additional smaller models.~\cite{kung2023active} proposes a new framework based on instantaneous uncertainty, which identifies informative tasks by measuring changes in the probability distribution of the model's outputs. Superfiltering~\cite{li2024superfiltering} utilizes smaller language models to filter out better data. By using GPT-2 combined with Perplexity and IFD~\cite{li2023quantity} to filter the data.

\paragraph{Data attribution}
The work of data attribution aims to quantify the contribution of each training sample to the model's prediction results. It attempts to answer the question, "Which training samples have the most significant impact on a specific prediction result?" In recent studies, ~\cite{li2023quantity} quantifies the degree of assistance provided by instructions to models by comparing the difficulty of generating responses with and without the instruction context. Nuggets~\cite{li2023one} leverage the downstream task benefits brought by data to the model for data filtering. LESS~\cite{xia2024less} estimates the impact of data by using gradients in the optimizer and performs low-rank gradient similarity searches for instruction data selection.

\paragraph{Knowledge in LLM}
From the perspective of model interpretability, it has been consistently observed that language models based on the transformer architecture encode "lower-level" information in the earlier layers and more "semantic" information in the later layers~\citep{tenney2019bert}.
Researches on the content knowledge of models~ \cite{meng2022locating,li2024inference} have emphasized the role of intermediate and top layers in factual prediction, as well as specific heads in ensuring truthfulness. DoLA~\cite{chuang2023dola} utilizes the fact-oriented characteristics of the top layers in LLMs to propose contrastive decoding, aiming to mitigate the hallucination problem during the generation process.

\vspace{1.0\baselineskip}

All these methods have their certain flaws. \textbf{Instruction data selection} methods still introduce biases from the auxiliary models used in the selection process though they have reduce the need for human intervention. \textbf{Data attribution} often require complex computations or additional fine-tuning steps. Building on these insights, our work addresses these limitations by proposing a novel approach that directly analyzes the weight differences between the fine-tuned and original models. By focusing on the weight differences in these layers, we can more effectively identify data that contributes to meaningful semantic changes in the model, rather than surface-level or lower-order modifications. This approach allows us to select data that is more likely to improve the model’s overall performance and generalization capabilities.

\input{latex/mxd/method}

\section{Experiments}
\subsection{Experiments Setup}

\paragraph{Training datasets} We conducted experiments separately on datasets in the following three different domains: general domain, code domain, and mathematics domain. In the general domain, two datasets DOLLY~\cite{conover2023free} and OPEN ASSISTANT~\cite{kopf2024openassistant} are included. In the domains of coding and mathematics, we use dataset evol-codealpaca-v1~\cite{luo2023wizardcoder} and dataset MetaMath~\cite{yu2023metamath} respectively. The detailed data structure format can be referred to in Appendix ~\ref{appendix_training}.

\paragraph{Evaluation} We follow the evaluation methodology of Open Instruct~\cite{wang2023far}, where MMLU~\cite{hendrycks2020measuring} and BBH~\cite{suzgun2022challenging} are used to assess the model's performance in general domains. We used GSM8k~\cite{cobbe2021training} to evaluate the model's performance in mathematical reasoning. For the code evaluation, we employed the HumanEval dataset~\cite{chen2021evaluating} to assess the model's ability to generate functionally correct programs from docstrings.In accordance with the settings of Open Instruct~\cite{wang2023far}, we refer to this dataset as HumanEval-CodeX.
For all experimental tasks, we used the greedy decoding method to obtain the generated results from the model.

\paragraph{Baselines} 
We compare ResoFilter with several baseline methods. In the experiments, we applied each data filtering method to produce three different amounts of training data for model training, specifically 25\%, 50\%, and 75\% of the total data. The simplest method is random, which randomly samples different amounts of training data from the dataset. For the Loss and PPL methods, we first ranked the data from high to low based on loss and PPL, and then selected the top 25\%, 50\%, and 75\% for experiments. Similarly, for Superfiltering~ \cite{li2024superfiltering} and Nuggets~ \cite{li2023one}, we filtered the data according to their respective methods, resulting in different amounts of training data. We used the last 3 layers diff data and took mean value of them as the filter score.
\input{latex/tza/table/combined_results}

\paragraph{Models and Training Parameters}
\input{latex/tza/table/exp_setting}
We conducted experiments using four models: Llama2-7B-Base~\cite{touvron2023llama}, Gemma2-2B-Base~\cite{team2024gemma}, Mistral-7B-Base-v0.3~\cite{mistral_7b_v0_3}, and Phi2-3.8B-Base-~\cite{javaheripi2023phi}. In the general domain, experiments were carried out with model Llama2-7B-Base. For the code domain, model Gemma2-2B-Base was utilized. In the mathematics domain, all four models were tested. We standardized the training hyperparameters, with the learning rate set to 1e-5 and using the AdamW optimizer. Other training parameters are shown in Table~\ref{exp_setting}.

\subsection{Main Results}
\paragraph{Method Comparision}

To evaluate the effectiveness and transferability of our method across different model architectures, we conducted experiments on three models: Gemma2-2B, Llama2-7B, and Llama2-13B. As shown in Table~\ref{combined_results}, our method consistently outperforms other selection approaches across all models. Notably, our method gets the best performance in 25\% among all the methods in all models, and for Gemma2-2B and Llama2-7B, our method achieves superior performance to full SFT at both 50\% and 75\%, demonstrating its ability to maintain high performance with reduced data. Moreover, our approach incurs minimal additional time cost compared to other methods, making it both effective and efficient. These results highlight the robustness and adaptability of our weight difference-based data selection strategy across various model sizes and architectures.

\paragraph{Generalizability}
To assess the generalizability of our proposed method, we conducted experiments across various domains, including general knowledge (MMLU), code generation (HumanEval-CodeX), and reasoning tasks (BBH). Table \ref{combined_results} presents the results of these cross-domain evaluations.

Our method consistently outperforms random sampling across different domains and model architectures, demonstrating its transferability. Notably, in the code domain (HumanEval-CodeX), our approach achieves significant improvements over random sampling, particularly at lower percentages of data selection (p25 and p50).

It's worth noting that for MMLU, both our method and random sampling show a decrease in performance compared to the base model. This is likely due to the base model's strong initial capabilities in general knowledge tasks, combined with the potential mismatch or lower quality of the fine-tuning data. As detailed in Appendix~\ref{appendix_training}, this phenomenon highlights the importance of data quality in fine-tuning, especially for models with high baseline performance in specific domains.

\paragraph{Scalability to Larger Models}
\input{latex/mxd/tables/model_scale}
To specifically validate our method's scalability, we conducted extended experiments on Gemma2-9B and Llama2-70B. Using the identical data filtered by their smaller counterparts (Gemma2-2B and Llama2-7B respectively), these larger models achieve comparable relative improvements to smaller models (3.2\% vs. 3.4\% average gain at p25). This demonstrates that: 1) The data selection criteria learned from smaller models generalize effectively to larger architectures; 2) Our method scales naturally with model capacity without requiring additional filtering iterations. The consistent performance across orders-of-magnitude parameter differences (2B to 70B) suggests our approach captures fundamental data quality characteristics rather than model-specific artifacts.


\section{Discussion}
\subsection{Alabtion}
To thoroughly investigate the robustness and effectiveness of our proposed method, we conduct a series of ablation studies. These experiments aim to provide a comprehensive understanding of various factors influencing the performance of our approach. Unless otherwise specified, we use the mean difference of weights in the last three layers as our default method for the following experiments. We explore the effects of different modules, layer indexes, statistical methods, training data number, and data ordering on our method's performance. The following subsections detail each of these aspects, offering insights into the key components and sensitivities of our approach.

\paragraph{Statistical Methods}
\input{latex/tza/table/stat}
The results in Table~\ref{stat_method} show that different statistical methods perform differently under different data ratios. At 25\%, the p99 percentile method performs the best; At 50\% and 75\%, the p90 percentile method achieved the best results, significantly better than the random baseline. The mean method performs stably at all data ratios and consistently outperforms random selection. In contrast, the performance of cosine similarity and Pearson correlation coefficient methods is relatively weak. These findings indicate that selecting appropriate statistical methods is crucial for effective data screening, especially percentile and mean methods, which demonstrate strong robustness and effectiveness in this task.

\paragraph{Layer Position}
\input{latex/tza/figures/layer_idx}
As shown in Figure~\ref{layer_idx} showed that on 25\% of the data, the model performance showed a significant trend with the change of layers. Shallow (1-8 layers) filtered data resulted in lower GSM8k scores, while deep (20-26 layers) filtered data produced higher scores. This indicates that data with significant differences in deep weights may better capture features that affect task performance. It is worth noting that the 25\% curve shows a clear upward trend, while the 75\%  curve is relatively stable, indicating that for 75\%, there is not much difference between deep screening and shallow screening. This difference implies changes in data sensitivity at different levels, providing some reference for optimizing data selection strategies.

\paragraph{Train Dataset Number}
\input{latex/tza/figures/train_num}
The results in Figure~\ref{train_num} show that as the number of train dataset increases, the scores of both methods (Diff and Random) show an upward trend. Prior to P25, Random is higher than our method, demonstrating that diversity is more important than quality when the data volume is small. Subsequently, the two curves intersected, and the performance of the Diff method(Ours) was generally better than that of the Random method. This indicates that when there is basic diversity, using weight differences to select higher quality data can help achieve better results. As the amount of training data increases, the gap between the two lines also widens, eventually reaching its maximum value (around~3 points) at p90 and p95.


\subsection{Feature Analysis}
To better understand the characteristics of data samples that lead to different fine-tuning outcomes, we performed a comprehensive analysis of the High-Diff Value, Low-Diff Value, and Random Sample datasets. Our analysis focused on the Gemma2-2b model, utilizing the mean difference of the last three weight layers as the selection criterion. The High Diff Value set i.e "dirty sample" is the top 1\% of samples with the highest mean difference, while the Low Diff Value set i.e "good sample" means the bottom 1\% with the lowest mean difference. For comparison, we also included a Random Sample set. We examned these datasets across four key dimensions as shown in the following paragraphs. This analysis aims to uncover the underlying patterns that distinguish high-performing from low-performing fine-tuning data.

\paragraph{Token Lengths Distribution}
\input{latex/tza/figures/token_len}
As shown in the Figure~\ref{token_len}, the token length distribution of the High Diff Value dataset is significantly shorter where concentrated on the left side of the chart, indicating that these samples usually contain fewer tokens. In contrast, the distribution of the Low Diff Value dataset is more dispersed and tends towards longer token sequences, with its peak appearing in the middle right part. Random samples exhibit a relatively uniform distribution, mainly concentrated in the medium length range.

\paragraph{Token Frequency}
\input{latex/tza/figures/token_freq}
As shown in the Figure~\ref{token_freq}, the High Diff Value data exhibits significant peaks in the low-frequency region, indicating that this type of sample tends to use rare or special vocabulary. In contrast, the distribution of Low Diff Value data is more uniform, covering a wider frequency range, suggesting that these samples use more common vocabulary. Random Sample falls between the two and presents a relatively balanced distribution.

\paragraph{Unique Token Ratio}
\input{latex/tza/figures/uniq_token}
As shown in the Figure~\ref{uniq_token}, the three types of data exhibit significantly different distribution characteristics. The unique token ratio of Low Diff Value samples is concentrated in a lower range (0.1-0.3), indicating that these samples tend to use more repetitive vocabulary. In contrast, the distribution of High Diff Value samples tends towards higher ratios (0.4-0.7), indicating that these samples contain more unique vocabulary. Random Sample falls between the two, with a relatively uniform distribution.

\paragraph{Cosine Similarity Distribution}
\input{latex/tza/figures/cos_sim}
As shown in the Figure~\ref{query_sim}, the similarity distribution of the three types of data shows significant differences. The distribution of Low Diff Value data tends to the right, indicating a high degree of query similarity within this type of data, concentrated in the range of 0.45-0.55. In contrast, the distribution of High Diff Value data is skewed towards the left, mainly in the range of 0.35-0.45, indicating a lower internal query similarity. The distribution of random samples falls between the two. This result indicates that Low Diff Value data has higher internal consistency, while High Diff Value data tends to be more diversity.

\paragraph{Analysis}
Based on the analysis of the above four aspects, we can summarize the significant feature differences between Low Diff Value data and High Diff Value data:

Low Diff Value data tends to repeatedly use more common vocabulary to form longer token sequences, and has high internal consistency. In contrast, High Diff Value data contains more rare or special vocabulary and refuses to expand sequence length, resulting in higher diversity. These feature differences may explain why Low Diff Value data is crucial for model training performance while High Diff Value data becomes 'dirty data'. The uniqueness and inconsistency of High Diff Value data may introduce noise/contain outliers/incorrect labeling/information unrelated to the main task, thereby interfering with the model's learning of general patterns. On the contrary, the consistency and universality of Low Diff Value data help the model capture key features and common patterns, maintain model stability, avoid overfitting to small or abnormal samples, and thus improve generalization ability and overall performance.



\input{latex/mxd/con_limi}

\bibliography{custom}

\appendix

\section{Training Data Format}
\label{appendix_training}
When constructing the instruction fine-tuning data, we followed the formats: Gemma2, Phi2, and Mistral used one set of templates, while Llama2 used another set of templates. Taking the Math dataset as an example, the specific formats are as follows:
\begin{tcolorbox}[colback=gray!5!white,colframe=gray!75!black,title={Gemma2, Phi2, Mistral}]
\textbf{<start\_of\_turn>}user \\ 
Calculate 8 divided by $\frac{1}{8}$.\textbf{<end\_of\_turn>}\\
\textbf{<start\_of\_turn>}model\\
Dividing by a fraction is the same as multiplying by its reciprocal.\\
So, $8 \div \frac{1}{8} = 8 \times \frac{8}{1} = 64$.\\
The answer is: 64
\textbf{<end\_of\_turn>}
\end{tcolorbox}

\begin{tcolorbox}[colback=gray!5!white,colframe=gray!75!black,title=Llama2]
\textbf{[INST]}Dave bought 8 books about animals, 6 books about outer space, and 3 books about trains to keep him busy over the holidays. Each book cost $6$. How much did Dave spend on the books?.\textbf{[/INST]}Dave bought a total of 8 + 6 + 3 = 17 books\\
Each book cost $6$, so Dave spent a total of 17 x $6$ = $102$ on the books.\\
\#\#\#\# 102 \\
The answer is: 102
\textbf{</s>}
\end{tcolorbox}

\section{Analysis of General Performance}
\label{mmlu_analysis}

To thoroughly investigate the performance on general tasks, we conducted comprehensive experiments across various training datasets and fine-tuning techniques. Our analysis focuses on two aspects: the impact of different training datasets and the effectiveness of different fine-tuning methods.

\subsection{Impact of Training Datasets and Fine-tuning Methods}
\input{latex/mxd/tables/mmlu_app}

For the MMLU benchmark, we performed experiments with several training sets, as detailed in Table~\ref{tab:mmlu_app}. The results show that it was nearly impossible to match the performance of the instruction-tuned version of the Gemma2 model (0.53). Moreover, we observed consistent performance degradation after fine-tuning compared to the base model.

To further understand this phenomenon, we conducted additional experiments comparing different fine-tuning approaches:

\input{latex/mxd/tables/mmlu_lora }
As shown in Table~\ref{tab:mmlu_finetuning}, both SFT and LoRA techniques demonstrate similar patterns of performance decline, suggesting this is a fundamental challenge in the field of model fine-tuning rather than a limitation of specific training approaches.

\subsection{Performance on BBH}
\input{latex/mxd/tables/bbh_app}

Our experiments on BBH revealed an interesting pattern. As shown in Table~\ref{tab:bbh_app}, reducing the size of the training set often led to models that outperformed those fine-tuned on the full dataset. In certain datasets, smaller training sets even produced superior training outcomes.

This observation aligns with the findings of \citet{sun2024amurocharanalyzing}, who noted that general model capabilities are inherently challenging to enhance through fine-tuning. Furthermore, with respect to the Gemma2-2B model, it has exhibited exceptional performance given the limitations of its parameter scale, approaching the theoretical performance ceiling for models of similar size.

These results collectively suggest that the challenge in improving performance on general tasks like MMLU lies not in the specific training method or data selection approach, but rather in the fundamental limitations of current fine-tuning paradigms for enhancing general knowledge capabilities.

\section{Reproducibility Analysis}

To ensure the statistical reliability of our results, we conducted three independent runs with different random seeds while maintaining consistent experimental settings. All experiments were performed under identical environmental variables, training code and parameters, evaluation protocols, and data ordering (maintaining original order after filtering).

\input{latex/mxd/tables/reproducibility}

The results show that our method maintains stable performance across different runs, with score variations typically within 0.01 points. This demonstrates both the reliability of our method and the significance of the performance improvements over the baseline.

Through these comprehensive experiments, we have demonstrated several key strengths of ResoFilter. The method shows excellent scalability to larger models through efficient cross-model data selection, while maintaining consistent effectiveness across different model scales. The reproducibility analysis confirms that our approach is statistically reliable with stable results. Moreover, the successful application of data filtered by smaller models to larger ones validates the practical efficiency of our "filter-once-apply-many" approach.

\section{Module Type}
\input{latex/tza/table/module}
In most experiments we only use the $W_{up}$ of the weight parameters in model to validate our ideas. Would the self attention weights contain more infomation to help to select data? The results in Table~\ref{module} showed that the data selected based on the $W_{up}$ module achieved the best performance at all ratios, especially at data volumes of 50\% and 75\%, which were significantly better than the random baseline. Other modules, such as $W_{down}$ and $W_k$, also perform well under certain data ratios. In contrast, the $W_v$ module performs the worst at 25\% data volume, but all the modules follow the same trends.

\section{Data Order}
\input{latex/tza/table/order}
Inspired by the concept of curriculum learning, we explored the impact of data order on model performance. Table~\ref{order} presents the results of this experiment on the GSM8k benchmark. Interestingly, the original data order consistently outperformed the other arrangements across all percentages. The random order showed competitive performance, particularly at higher data percentages. The Min to Max order performed slightly better than Max to Min, especially at lower data percentages, suggesting that introducing easier examples first might be beneficial. However, the differences between these orderings were relatively small, indicating that the original data order already possesses an inherent structure that is conducive to effective learning.

\section{Algorithm}

First, for each sample \(d_i\) in the dataset \(D\), we fine-tune the initial model \(M_0\) using \(d_i\) to obtain \(M_i\). We then compute the parameter difference \(\Delta W = M_i - M_0\), and for each layer and module in \(M_i\), we calculate the average parameter difference.
After conducting a series of experiments, we selected the $W_{up}$ modules from the final n layers and computed their average differences, denoted as \( \text{diff}_i \). This allows us to associate \( \text{diff}_i \) with the corresponding sample \( d_i \).
Next, we perform data filtering by sorting the dataset \( D \) in descending order based on \( \text{diff}_i \). According to the desired number of retained samples \( k \), we select the bottom \( k \) samples from the sorted dataset \( D \) to form \( D_{\text{filtered}} \), while preserving their original relative order.
The filtered dataset \( D_{\text{filtered}} \) is then provided directly to the model for the fine-tuning process.
\input{latex/mxd/algorithm}

\end{document}

%% file: latex/mxd/abs.tex
\begin{abstract}
Large language models (LLMs) have shown remarkable effectiveness across various domains, with data augmentation methods utilizing GPT for synthetic data generation becoming prevalent. However, the quality and utility of augmented data remain questionable, and current methods lack clear metrics for evaluating data characteristics. To address these challenges, we propose \textbf{ResoFilter}, a novel method that integrates models, data, and tasks to refine datasets. ResoFilter leverages the fine-tuning process to obtain Data-Parameter features for data selection, offering improved interpretability by representing data characteristics through model weights. Our experiments demonstrate that ResoFilter achieves comparable results to full-scale fine-tuning using only \textbf{half the data} in mathematical tasks and exhibits strong generalization across \textbf{different models and domains}. This method provides valuable insights for constructing synthetic datasets and evaluating high-quality data, offering a promising solution for enhancing data augmentation techniques and improving training dataset quality for LLMs. For reproducibility, we will release our code and data upon acceptance.
The source code and implementation details of this work are publicly available in our GitHub repository (\href{https://github.com/TAL-auroraX/ResoFilter}{https://github.com/TAL-auroraX/ResoFilter})
\end{abstract}

%% file: latex/mxd/intro.tex
\section{Introduction}

Large language models (LLMs) have demonstrated remarkable capabilities across various domains, with training data playing a pivotal role in enhancing their performance. The quality and quantity of data are crucial factors in all stages of LLM development, including pretraining, instruction tuning, and alignment~\citep{peters-etal-2018-deep,Radford2018ImprovingLU,devlin-etal-2019-bert,Raffel2019ExploringTL,touvron2023llama,mishra2021cross,sanh2022multitask,longpre2023flan,muennighoff2024generative,ziegler2020finetuning,bai2022constitutional,ouyang2022traininglanguagemodelsfollow,rafailov2023direct}.

The significance of high-quality datasets has been well-established in the pretraining phase, where data cleaning techniques have shown to substantially improve model performance, especially for smaller models~\citep{NEURIPS2019_c04c19c2,Raffel2019ExploringTL,wenzek-etal-2020-ccnet,gao2020pile,rae2022scaling,lee-etal-2022-deduplicating}. Following the success of ChatGPT~\cite{ouyang2022traininglanguagemodelsfollow,openai2024gpt4technicalreport}, the focus has shifted towards creating high-quality fine-tuning datasets, leading to a surge in methods for automatic generation of instruction-following data using GPT models.

While these methods, such as Self-Instruct~\citep{wang-etal-2023-self-instruct}, Evol-Instruct~\citep{xu2023wizardlm}, and others~\citep{honovich-etal-2023-unnatural,auggpt,abdullin2024synthetic}, have significantly increased the scale of available datasets, the notion of "high quality" remains contentious. As observed by ~\citet{schimanski2024faithful}, current approaches predominantly focus on augmenting data volume rather than enhancing quality. This trend has led to the phenomenon of "diminishing returns," where performance gains plateau as dataset size increases beyond a certain threshold, emphasizing the need to prioritize data quality over quantity.

\input{latex/qty/figures/pipeline1}

To address this challenge, we propose ResoFilter, a novel method that leverages the fine-tuning process for effective data selection. ResoFilter processes each data point through full forward and backward propagation, capturing the induced change in model weights. From these changes, we derive a characteristic score for each data point, which serves as the metric for subsequent selection. Our approach offers improved interpretability by representing data characteristics through model weights, building upon existing research demonstrating that model weights store knowledge~\cite{hanna2023doesgpt2computegreaterthan,dai-etal-2022-knowledge}. Figure~\ref{pipeline1} illustrates the detailed workflow of our method.

Empirical results demonstrate the efficacy of ResoFilter. In the context of MetaMath~\cite{yu2023metamath}, we achieve comparable performance to full dataset fine-tuning using only 50\% of the data selected by our method. Moreover, by eliminating poorly performing data points, ResoFilter can even surpass the performance of full fine-tuning. The method also exhibits strong generalization across different models and domains, including mathematics, code, and general question answering tasks.

Our contributions can be summarized as follows:

\begin{enumerate}
    \item We introduce ResoFilter, an effective method for selecting high-quality datasets from large-scale data collections to enhance the fine-tuning process of large language models.
    
    \item We validate the generalizability of ResoFilter by demonstrating its excellent performance across various domains (including mathematics, code, and general tasks) and different model architectures, showcasing its robustness and wide applicability.
    
    \item Through extensive experimental analysis, we provide valuable insights into the construction of synthetic datasets and the analysis of high-quality data, offering guidance for future data synthesis and selection methodologies.
\end{enumerate}

%% file: latex/qty/figures/pipeline1.tex

\begin{figure*}[!t]
  \centering
  \resizebox{0.85\textwidth}{!}{
    \includegraphics{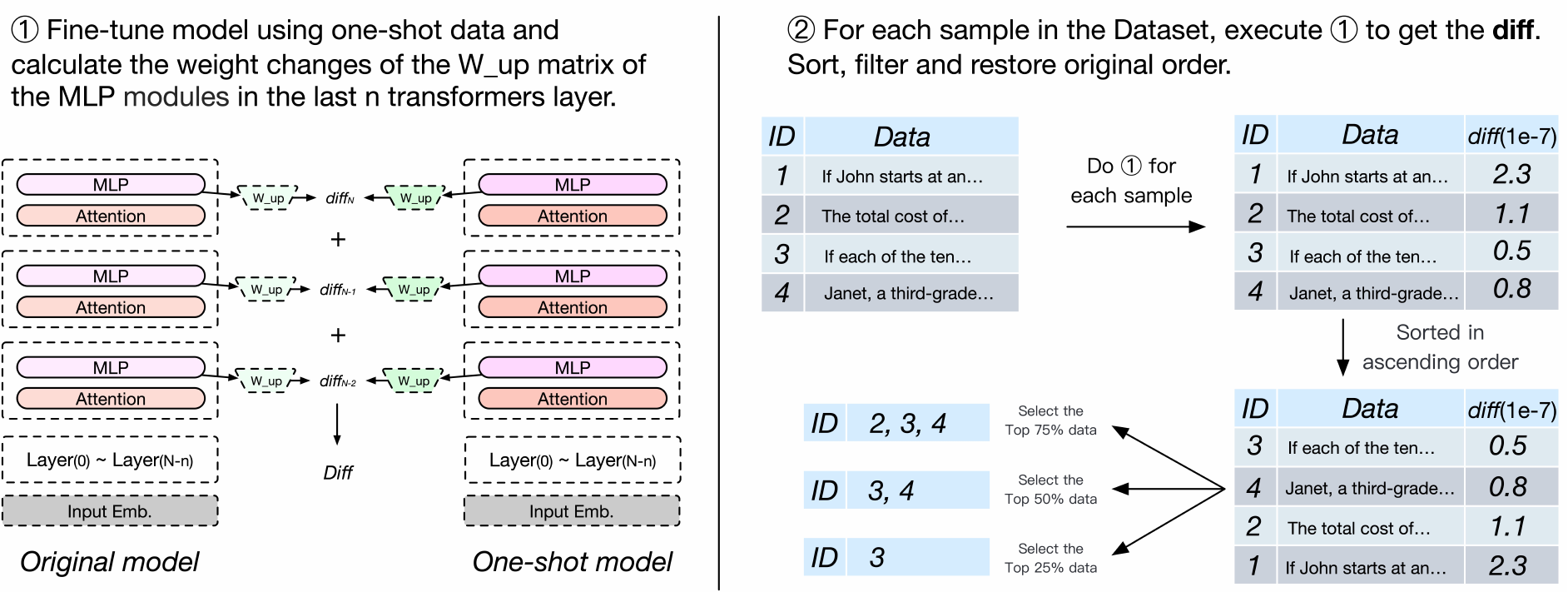}
  }
  \vspace{-5pt}
  \caption{Workflow for our method. The left side of the figure illustrates the detailed process of calculating parameter changes for individual data samples using the $W_{up}$ matrix from the last n layers of the neural network. The right side demonstrates the application of this method to the entire dataset, including steps for computing parameter change values for each sample, sorting based on these values, filtering out samples with the largest parameter changes, and restoring the remaining data to its original order.}
  \label{pipeline1}
  \vspace{1pt}
\end{figure*}

%% file: latex/mxd/method.tex
\section{Method}

\subsection{Background and Rationale}

Recent advancements in data selection methods focus on improving data representation and effectively combining these representations for data selection \cite{Take_essence}. While some approaches leverage GPT-4 scoring directly \cite{cao2023instruction,wei2023instructiongpt4}, we argue for methods that integrate diverse model features for enhanced interpretability \cite{xia2024less}.

Research on model interpretability has revealed the importance of MLP layers and deep layers in feature learning \cite{hanna2023doesgpt2computegreaterthan,dai-etal-2022-knowledge}. Additionally, studies on model editing \cite{edit_CorrectingLM} and distillation \cite{Yao2024ScalableME} have explored knowledge distribution across various network layers. Building on these insights, we propose a framework that integrates the model, data, and task.

While our work shares some conceptual similarities with gradient-based methods like LESS \cite{xia2024less}, our approach differs in two fundamental aspects:

\begin{enumerate}
    \item \textbf{Scope and Nature of Information:} While gradient-based methods focus on immediate output-level changes through backpropagation, our method captures holistic parameter changes across the entire model during fine-tuning. Gradients represent instantaneous directional changes at a specific point in parameter space, whereas our method observes actual parameter updates and their stabilized effects after optimization.
    
    \item \textbf{Interpretability and Causal Understanding:} Our work is influenced by research on model interpretability \cite{hanna2023doesgpt2computegreaterthan}, aiming to understand the causal chain between data characteristics and model behavior. By analyzing the relationship between parameter changes, data characteristics, and downstream task performance, we provide insights into how different types of data influence model behavior - a three-way analysis that pure gradient-based approaches cannot capture.
\end{enumerate}

These distinctions make our method particularly suitable for analyzing and selecting high-quality instruction tuning data, where understanding the comprehensive and longer-term effects of data on model behavior is crucial.

\subsection{Problem Definition and Modeling}

In the context of Supervised Fine-Tuning (SFT), we aim to balance data richness and its impact on the model for specific tasks. We formulate this as an optimization problem with the following objective function:

\begin{multline}
    E(D, M, p) = F_{richness}(p) \cdot \\
    (1 + \beta \cdot F_{characteristic}(p))
\end{multline}

where \(D\) is the dataset, \(M\) is the model, and \(p\) is the filtering percentage.

\paragraph{Objective Function Design}
The multiplicative form ensures interdependence between data richness and characteristic intensity. The term \((1 + \beta \cdot F_{characteristic})\) provides a baseline guarantee, ensuring that data richness is always considered even when characteristic intensity is low. This design captures the non-linear relationship between data quantity and quality, which is more realistic than a simple linear combination.

As the filtering percentage \(p\) increases, \(F_{richness}(p)\) grows, representing the benefit of including more data, while \(F_{characteristic}(p)\) captures the relevance of the selected data subset to the specific model and task.

The parameter \(\beta\) modulates the influence of characteristic intensity relative to data richness, while \(\lambda\) in \(F_{richness}(p) = 1 - e^{-\lambda |D_p|}\) controls how quickly data richness approaches its maximum value. These parameters enable our problem formulation to adapt to diverse datasets and fine-tuning scenarios within the SFT framework.

Importantly, this objective function captures a crucial trade-off: as \(p\) increases, \(E\) reaches an optimal point, balancing data quantity and quality. Beyond this point, although performance may decline, increased richness mitigates the impact of decreasing characteristic intensity, preventing catastrophic drops in effectiveness across larger data subset sizes.

\paragraph{Data Richness Modeling}
We model data richness using an exponential function:

\begin{equation}
    F_{richness}(p) = 1 - e^{-\lambda |D_p|}
\end{equation}

This function is bounded between 0 and 1, providing a normalized measure of richness. It increases monotonically as more data points are included, while exhibiting diminishing returns. This design reflects the real-world scenario where the marginal benefit of additional data decreases as the dataset grows, approaching a theoretical maximum richness as \(|D_p|\) tends to infinity.

\paragraph{Characteristic Intensity}
The characteristic intensity measures the contribution of the selected data subset to the model on a given task:

\begin{equation}
    F_{characteristic}(p) = \frac{1}{|D_p|} \sum_{j \in D_p} f(j, M)
\end{equation}

Here, \(f(j, M)\) quantifies the impact of data point \(j\) on the model \(M\). This function captures the unique characteristics of each data point in relation to the model.

\subsection{ResoFilter}

Our proposed method, ResoFilter, comprises two key components: a data screening process and a characteristic intensity calculation. The detailed algorithm is presented in Appendix D.

\paragraph{Data Screening Process}
We select a subset of data, \(D_p\), based on a ranking function:

\begin{multline}
    D_p = \{j \in D \mid \\
    \text{rank}(s(j, M)) > |D| \cdot (1 - p/100)\}
\end{multline}

Here, \(s(j, M)\) is a scoring function that ranks each data point \(j\) based on its relevance to the model \(M\). We consider data points causing smaller differences in the last n layers of the model as potentially more valuable, as they are less likely to disrupt previously acquired knowledge.

\paragraph{Characteristic Intensity Calculation}
The function \(f(j, M)\) evaluates the impact of each data point on the model's last n layers, ensuring selected data is both relevant and representative of key dataset features.

\paragraph{Module Selection}
Based on our comprehensive analysis (detailed in Appendix D), we identified $W_{up}$ as the most effective weight module for data filtering. As shown in Table~\ref{module}, $W_{up}$ consistently outperforms other modules across different data ratios, with particularly strong performance at 50\% and 75\% data volumes. This selection is further supported by ablation studies demonstrating the module's stability and effectiveness in capturing semantic changes during fine-tuning.

%% file: latex/tza/table/combined_results.tex
\begin{table*}[t!]
    \centering
    \resizebox{0.75\textwidth}{!}{
    \begin{tabular}{l|l|l|cccc|c}
    \toprule
    \textbf{Category} & \textbf{Model/Field} & \textbf{Method} & \textbf{25\%} & \textbf{50\%} & \textbf{75\%} & \textbf{Full SFT} & \textbf{Time\textsuperscript{*}} \\
    \midrule
    \multirow{18}{*}{Models} 
    & \multirow{6}{*}{Gemma2-2B} 
    & Random & 0.6042 & 0.6254 & 0.6474 & \multirow{6}{*}{0.649} & - \\
    & & Loss & 0.6034 & 0.6436 & 0.6436 & & 0.8 \\
    & & PPL & 0.6035 & 0.6467 & 0.655 & & 0.8 \\
    & & Superfiltering & 0.5079 & 0.6262 & 0.6482 & & 0.8 \\
    & & Nuggest & 0.5565 & 0.6164 & 0.6543 & & \textbf{98} \\
    & & Our method & \textbf{0.6042} & \textbf{0.6497} & \textbf{0.6603} & & 1.5 \\
    \cmidrule(lr){2-8}
    & \multirow{5}{*}{Llama2-7B} 
    & Random & 0.5525 & 0.5921 & 0.5981 & \multirow{5}{*}{0.5731} & - \\
    & & Loss & 0.5034 & 0.5413 & 0.5459 & & 1.7 \\
    & & PPL & 0.511 & 0.5353 & 0.5451 & & 1.7 \\
    & & Superfiltering & 0.3972 & 0.6073 & \textbf{0.6531} & & 2 \\
    & & Our method & \textbf{0.5549} & \textbf{0.6196} & 0.6444 & & 4.5 \\
    \cmidrule(lr){2-8}
    & \multirow{5}{*}{Llama2-13B} 
    & Random & 0.6276 & 0.6747 & 0.6686 & \multirow{5}{*}{0.6935} & - \\
    & & Loss & 0.6096 & 0.6209 & 0.6641 & & 4.9 \\
    & & PPL & 0.6141 & 0.6338 & 0.6611 & & 4.9 \\
    & & Superfiltering & 0.5496 & 0.677 & 0.6831 & & 5.5 \\
    & & Our method & \textbf{0.6322} & \textbf{0.6853} & \textbf{0.6855} & & 9 \\
    \midrule
    \multirow{9}{*}{Domains} 
    & \multirow{3}{*}{CodeX} 
    & Random & 0.2979 & 0.3549 & \textbf{0.404} & \multirow{3}{*}{0.4098} & \\
    & & Loss & 0.3146 & 0.3442 & 0.3872 & & \\
    & & Our method & \textbf{0.3579} & \textbf{0.425} & 0.3966 & & \\
    \cmidrule(lr){2-8}
    & \multirow{3}{*}{MMLU} 
    & Random & 0.5105 & \textbf{0.5152} & 0.4858 & \multirow{3}{*}{0.4658} & \\
    & & Loss & 0.507 & 0.501 & \textbf{0.4961} & & \\
    & & Our method & \textbf{0.5129} & 0.4893 & 0.4952 & & \\
    \cmidrule(lr){2-8}
    & \multirow{3}{*}{BBH} 
    & Random & 0.3796 & \textbf{0.3759} & \textbf{0.3824} & \multirow{3}{*}{0.3704} & \\
    & & Loss & 0.3796 & 0.3667 & 0.3685 & & \\
    & & Our method & \textbf{0.3861} & 0.362 & 0.3778 & & \\
    \bottomrule
    \end{tabular}}
    \caption{Combined results of cross-model and cross-domain evaluations. The upper section compares ResoFilter with baselines across model architectures (Gemma2-2B, Llama2-7B, Llama2-13B), while the lower section demonstrates domain generalization (CodeX, MMLU, BBH). \textit{Time\textsuperscript{*}} indicates training time cost (hours) per method, which is only applicable to model architecture experiments.}
    \label{combined_results}
\end{table*}

%% file: latex/tza/table/exp_setting.tex
\begin{table}[t]
\centering
\scalebox{0.85}{
\begin{tabular}{l|ccc}
\toprule	
& \textbf{MaxLegnth} & \textbf{BatchSize} & \textbf{Epoch} \\ 
\midrule
General Data     & 2048       & 16         & 3     \\
Math Data        & 768        & 64         & 1     \\
Code Data        & 2048       & 16         & 1    \\
\bottomrule
\end{tabular}
}
\caption{Detail parameter settings for model training.}
\label{exp_setting}
\end{table}

%% file: latex/mxd/tables/model_scale.tex
\begin{table}[htbp]
\centering
\resizebox{0.48\textwidth}{!}{
\begin{tabular}{lccccc}
\toprule
\textbf{Model} & \textbf{Base Score} & \textbf{Method} & \textbf{25\%} & \textbf{50\%} & \textbf{75\%} \\
\midrule
\multirow{2}{*}{Gemma2-9B} & \multirow{2}{*}{0.69} & Random & 0.7498 & 0.7467 & 0.7483 \\
 & & Ours & 0.7596 & 0.7566 & 0.7741 \\
\midrule
\multirow{2}{*}{Llama2-70B} & \multirow{2}{*}{0.61} & Random & 0.7671 & 0.7839 & 0.7960 \\
 & & Ours & 0.7733 & 0.7877 & 0.8059 \\
\bottomrule
\end{tabular}
}
\caption{Performance comparison on larger models using data filtered by smaller models (Gemma2-2B for Gemma2-9B, Llama2-7B for Llama2-70B).}
\label{tab:large_models}
\end{table}

%% file: latex/tza/table/stat.tex
\begin{table}[t!]
    \centering
    \resizebox{0.35\textwidth}{!}{
    \begin{tabular}{l|ccc}
    \toprule				
    \textbf{Methods} & \textbf{25\%}  & \textbf{50\%} & \textbf{75\%} \\
    \midrule
    \text{Random} & 0.6042 & 0.6254 & 0.6474 \\
    \cmidrule(lr){1-4}
    \text{Mean} & 0.6042 & 0.6497 & 0.6603 \\
    \text{Std} & 0.5812 & 0.6406 & 0.6459 \\
    \text{p90} & 0.5754 & \textbf{0.6504} & \textbf{0.6686} \\
    \text{p95} & 0.5928 & 0.6285 & 0.6542 \\
    \text{p99} & \textbf{0.6156} & 0.6474 & 0.6611 \\
    \text{cosine} & 0.5914 & 0.6179 & 0.6361 \\
    \text{pearson} & 0.5898 & 0.6391 & 0.6346 \\
    \bottomrule
    \end{tabular}}
    \caption{Here are the GSM8K scores applied various statistical indicators to the $W_{up}$ weight layer of the model, including 5 main statistical methods shown in the table. The p99 achieves the best performance in 25\% while p90 get the best in 50\% and 75\%.}
    \label{stat_method}
\end{table}

%% file: latex/tza/figures/layer_idx.tex
\begin{figure}[!t]
  \centering
  \begin{minipage}{0.478\textwidth}
   \centering
 \resizebox{1.\textwidth}{!}{
  \includegraphics{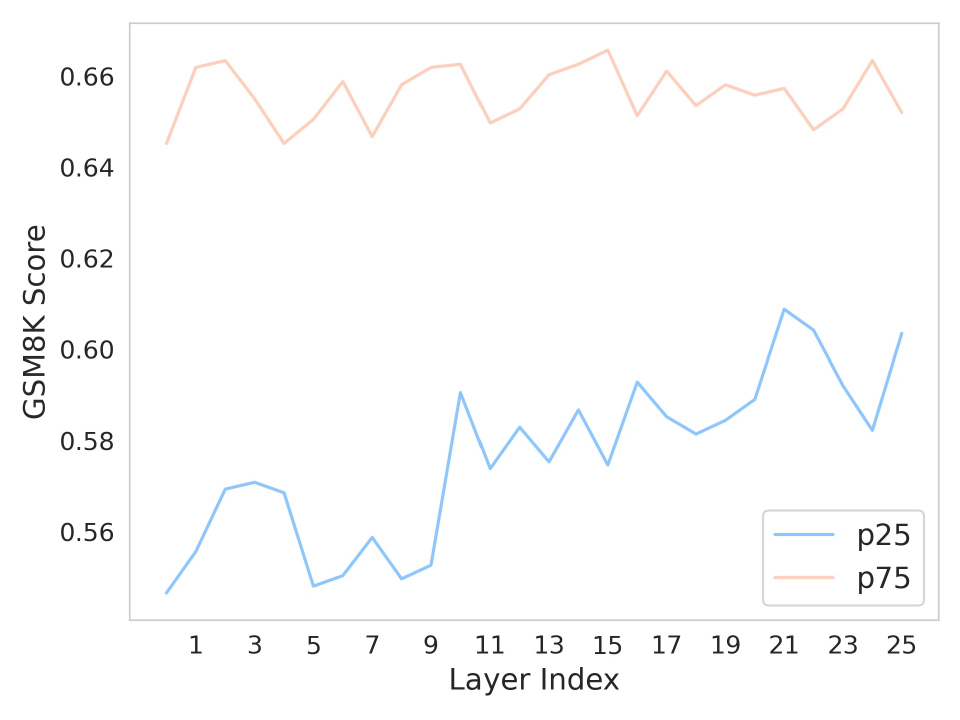}
  }
   \vspace{-5pt}
  \caption{We analyzed the $W_{up}$ weights of the model from the first layer to the 26th layer. The 25\% shows a continuous upward trend, while the 75\% fluctuates within a certain range.}
  \label{layer_idx}
  \end{minipage}
\end{figure}

%% file: latex/tza/figures/train_num.tex
\begin{figure}
  \centering
  \begin{minipage}{0.478\textwidth}
   \centering
 \resizebox{1.\textwidth}{!}{
  \includegraphics{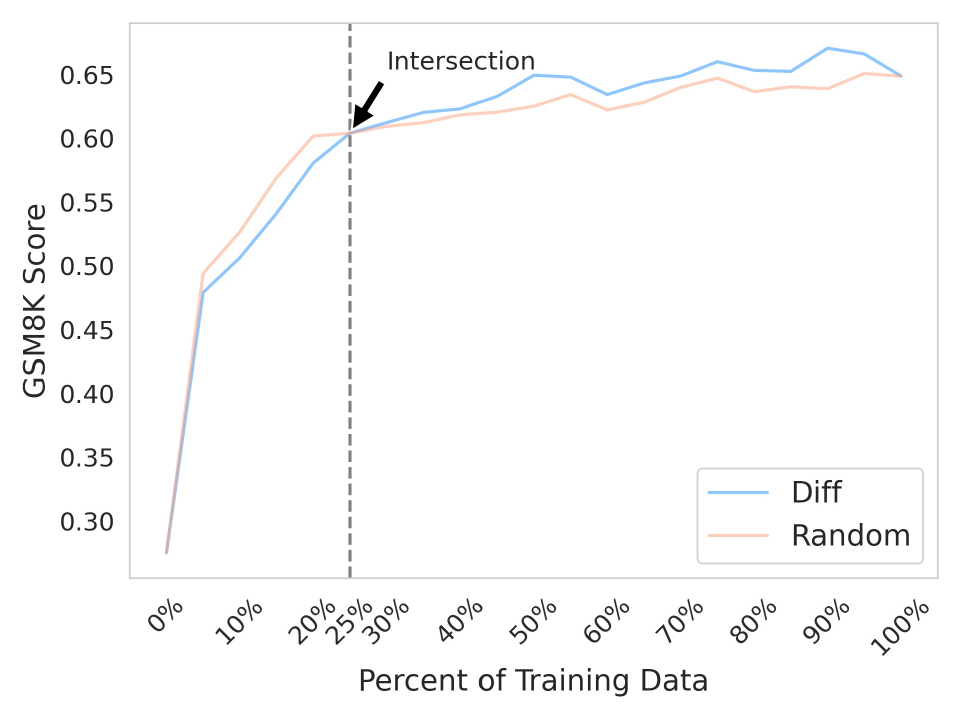}
  }
   \vspace{-5pt}
  \caption{Here we used filtering based on different proportions, and then trained and evaluated using the remaining data. The figure shows the GSM8K scores under different filtering ratios.}
  \label{train_num}
  \end{minipage}
\end{figure}

%% file: latex/tza/figures/token_len.tex
\begin{figure}[!t]
  \centering
  \begin{minipage}{0.478\textwidth}
   \centering
 \resizebox{1\textwidth}{!}{
  \includegraphics{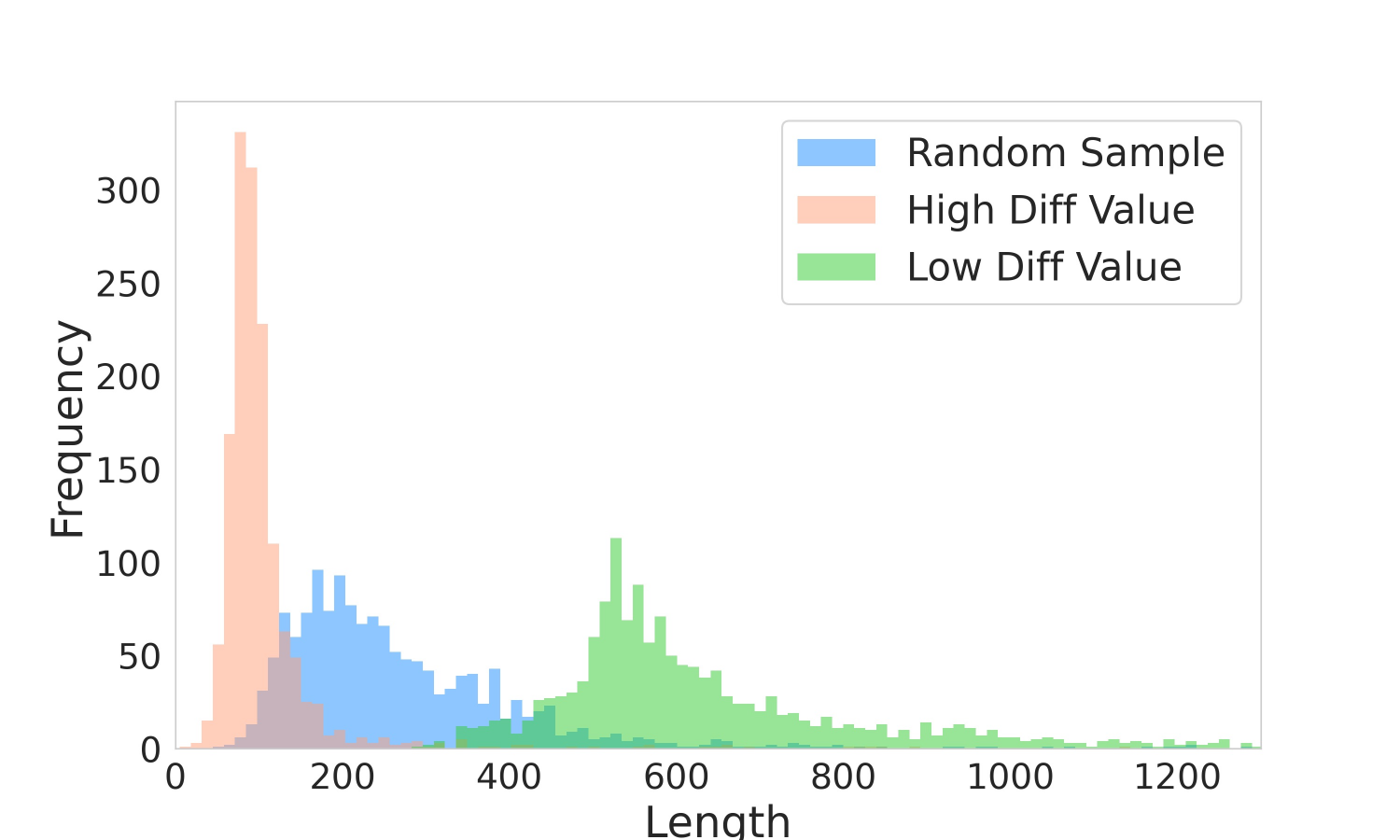}
  }
   \vspace{-5pt}
  \caption{Illustrates the token length distribution comparison across three datasets. Low Diff Value has much larger token numbers than High Diff Value data.}
  \label{token_len}
  \end{minipage}
\end{figure}

%% file: latex/tza/figures/token_freq.tex
\begin{figure}[!t]
  \centering
  \begin{minipage}{0.478\textwidth}
   \centering
 \resizebox{1.\textwidth}{!}{
  \includegraphics{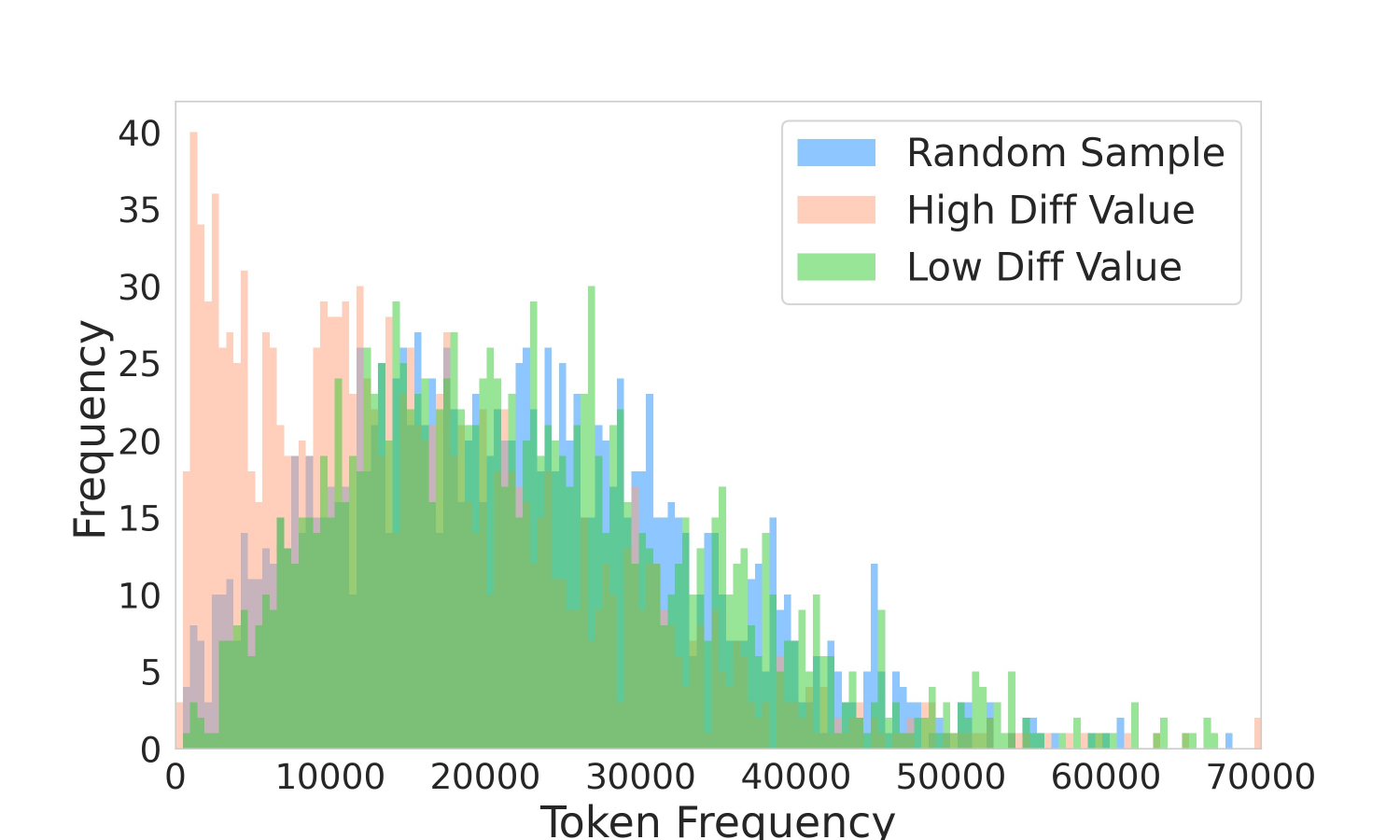}
  }
   \vspace{-5pt}
  \caption{Illustrates the token Frequency distribution comparison across three datasets. In order to analyze the frequency distribution of tokens, we first perform token statistics on the entire dataset to obtain the word frequency of each token. Subsequently, we calculated the total frequency of all token words in each sample divided by the sample length to obtain the average token frequency. High Diff Value shows more obscure vocabulary is used compared to Low Diff Value.}
  \label{token_freq}
  \end{minipage}
\end{figure}

%% file: latex/tza/figures/uniq_token.tex
\begin{figure}[!t]
  \centering
  \begin{minipage}{0.478\textwidth}
   \centering
 \resizebox{1.\textwidth}{!}{
  \includegraphics{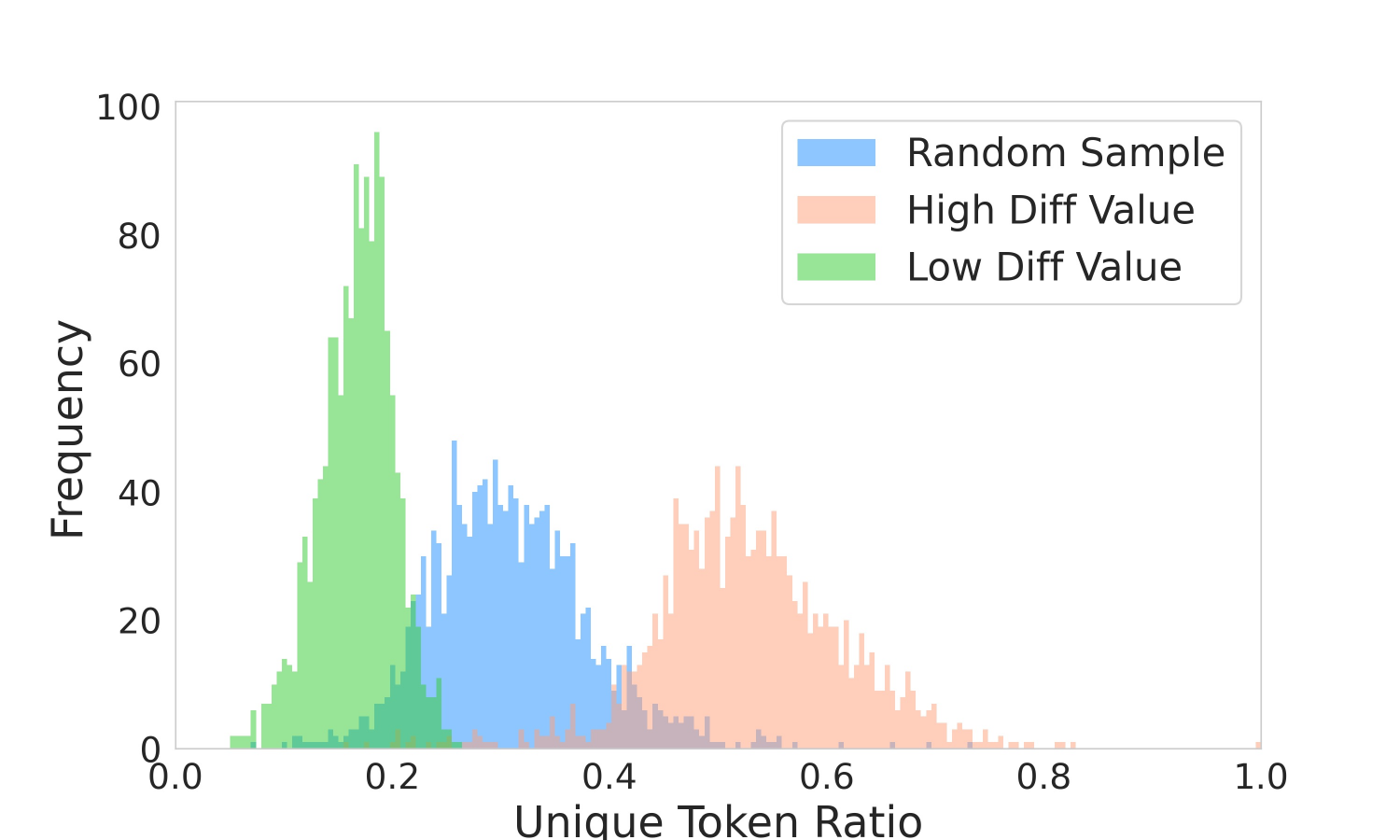}
  }
   \vspace{-5pt}
  \caption{To evaluate the lexical diversity of the dataset, we calculated the ratio of unique tokens in each sample, which is the number of unique tokens in each sample divided by the total number of tokens in the sample. Similar to Figure~\ref{token_len}, High Diff Value data shows every sample contains more unique words.}
  \label{uniq_token}
  \end{minipage}
\end{figure}

%% file: latex/tza/figures/cos_sim.tex
\begin{figure}[!t]
  \centering
  \begin{minipage}{0.478\textwidth}
   \centering
 \resizebox{1.\textwidth}{!}{
  \includegraphics{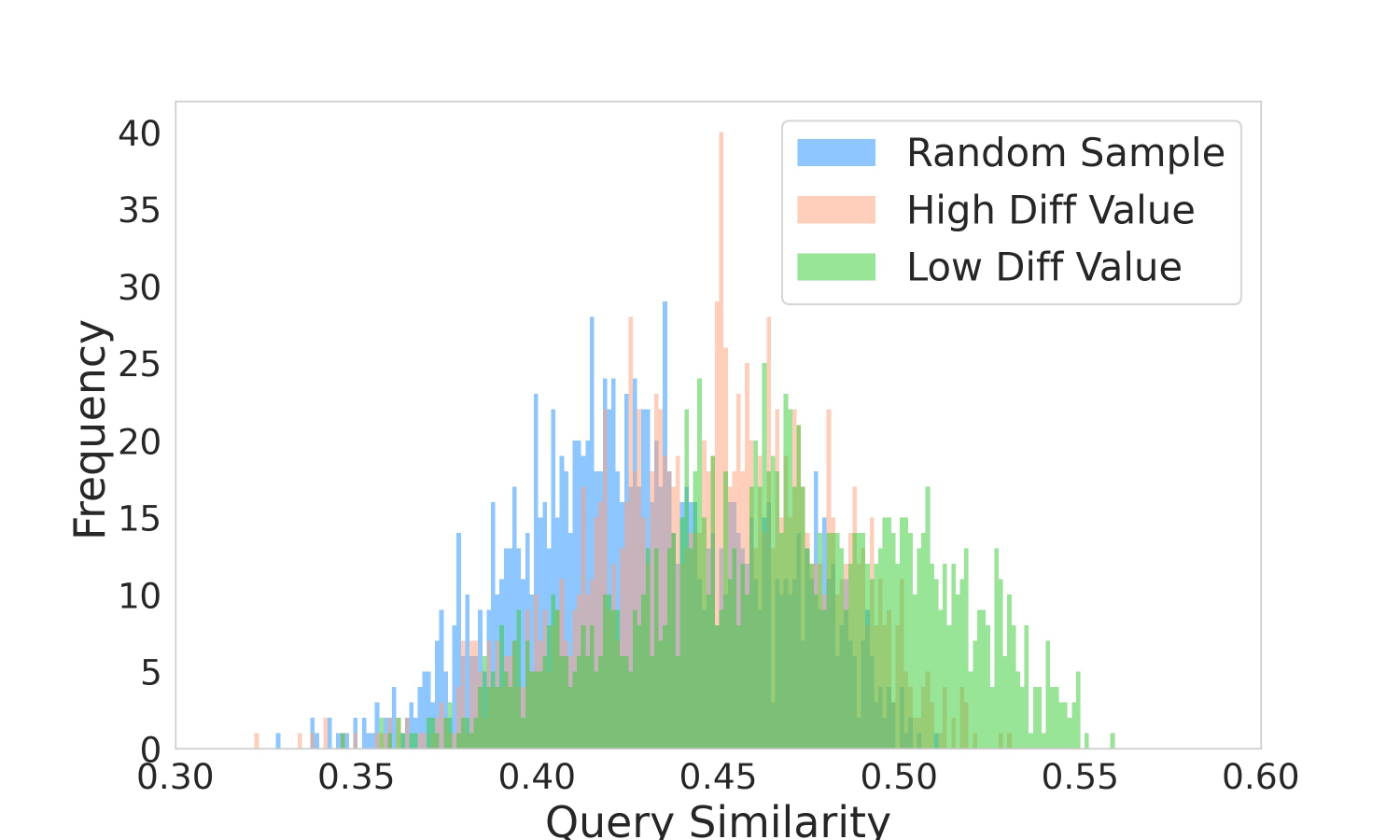}
  }
   \vspace{-5pt}
  \caption{In order to gain a deeper understanding of the characteristics of High Diff Value and Low Diff Value data, we analyzed the similarity distribution between texts. We use a pretrained embedding model~\cite{bge_m3} to calculate the cosine similarity between each query and the rest queries in the same class, and take the average. This obtained the average similarity distribution of three types of data. Low Diff Value data are much similar each other than else.}
  \label{query_sim}
  \end{minipage}
\end{figure}

%% file: latex/mxd/con_limi.tex
\section{Conclusion}

In this paper, we introduced ResoFilter, a novel approach for refining fine-tuning datasets obtained through GPT-based data augmentation. Our experiments showed that ResoFilter can maintain or improve training performance while removing up to 50\% of redundant data across various model architectures, sizes, and task domains. This work provides insights into balancing data richness and specificity, optimizing model performance and computational efficiency. ResoFilter enhances model training and opens new avenues for research in data annotation, augmentation, and domain-specific fine-tuning strategies, with potential applications in developing more robust and efficient language models.


\section*{Limitations}

Despite the demonstrated efficacy of ResoFilter through empirical experimentation, several limitations and avenues for future research persist. Firstly, the methodology could be further refined through the implementation of a multi-indicator approach, potentially enhancing the granularity and precision of the data filtering process. Secondly, the insights derived from ResoFilter could be utilized to inform the reverse-engineering of dataset construction, potentially yielding more optimal training datasets ab initio. Finally, the current study is constrained by the absence of comprehensive evaluations on very large-scale models (e.g., those exceeding 70 billion parameters), alternative architectures such as Mixture of Experts (MoE), and specialized systems like conversational models. Addressing these limitations in subsequent research would provide a more exhaustive understanding of ResoFilter's applicability and impact across a diverse spectrum of model scales and architectures.

%% file: latex/mxd/tables/mmlu_app.tex
\begin{table}[htbp]
\centering
\resizebox{0.45\textwidth}{!}{
\begin{tabular}{lc}
\toprule
\textbf{Model} & \textbf{MMLU Score} \\
\midrule
Gemma2-2b & 0.5300 \\
Gemma2-2b-it & 0.5605 \\
DOLLY \& OPEN ASSISTANT & 0.4658 \\
tulu-v1-sft-mixture & 0.3733 \\
mixed & 0.4087 \\
WizardLM  & 0.3184 \\
Flan V2 & 0.4064 \\
Flan V2 \& CoT & 0.4244 \\
\bottomrule
\end{tabular}
}
\caption{Comparison of different models on MMLU.}
\label{tab:mmlu_app}
\end{table}

%% file: latex/mxd/tables/mmlu_lora.tex
\begin{table}[htbp]
\centering
\resizebox{0.48\textwidth}{!}{
\begin{tabular}{lcc}
\toprule
\textbf{Training Dataset/Model} & \textbf{SFT} & \textbf{LoRA (rank=64)} \\
\midrule
NO Training (Base Model) & 0.5300 & - \\
Flan-v2 \& COT & 0.4244 & 0.495 \\
Dolly & 0.4658 & 0.518 \\
Tulu-v2 & 0.3733 & 0.494 \\
WizardLM & 0.3184 & 0.520 \\
\bottomrule
\end{tabular}
}
\caption{MMLU performance comparison across different training datasets and fine-tuning methods.}
\label{tab:mmlu_finetuning}
\end{table}

%% file: latex/mxd/tables/bbh_app.tex
\begin{table}[h]
\centering
\resizebox{0.5\textwidth}{!}{
\begin{tabular}{l|cccc}
\toprule
\textbf{Dataset} & \textbf{full} & \textbf{25\%} & \textbf{50\%} & \textbf{75\%} \\
\midrule
DOLLY \& OPEN ASSISTANT & 0.3704 & 0.3861 & 0.3620 & 0.3778 \\
Flan V2 \& CoT & 0.3148 & 0.3435 & 0.3056 & 0.2861 \\
MetaMath & 0.3565 & 0.3889 & 0.4000 & 0.3454 \\
evol-codealpaca-v1 & 0.3769 & 0.3861 & 0.3824 & 0.3870 \\
\bottomrule
\end{tabular}}
\caption{Our evaluation results on the BBH.}
\label{tab:bbh_app}
\end{table}

%% file: latex/mxd/tables/reproducibility.tex
\begin{table}[htbp]
\centering
\resizebox{0.45\textwidth}{!}{
\begin{tabular}{lccc}
\toprule
\textbf{Method} & \textbf{25\%} & \textbf{50\%} & \textbf{75\%} \\
\midrule
Ours-run1 & 0.6042 & 0.6497 & 0.6603 \\
Ours-run2 & 0.5998 & 0.6515 & 0.6586 \\
Ours-run3 & 0.6078 & 0.6454 & 0.6632 \\
\midrule
Ours-Avg & 0.6039 & 0.6489 & 0.6607 \\
Baseline(random) & 0.6042 & 0.6254 & 0.6474 \\
\bottomrule
\end{tabular}
}
\caption{Results from multiple experimental runs on Gemma2-2B, demonstrating the stability of our method.}
\label{tab:reproducibility}
\end{table}

%% file: latex/tza/table/module.tex
\begin{table}[t!]
    \centering
    \resizebox{0.4\textwidth}{!}{
    \begin{tabular}{l|ccc}
    \toprule				
    \textbf{Module} & \textbf{25\%}  & \textbf{50\%} & \textbf{75\%} \\
    \midrule
    \text{Random} & 0.6042 & 0.6254 & 0.6474 \\
    \cmidrule(lr){1-4}
    \text{$W_{up}$} & \textbf{0.6042} & \textbf{0.6497} & \textbf{0.6603} \\
    \text{$W_{down}$} & 0.589 & 0.652 & 0.6459 \\
    \text{$W_{q}$} & 0.5891 & 0.6338 & 0.649 \\
    \text{$W_{k}$} & 0.583 & 0.6444 & 0.6596 \\
    \text{$W_{v}$} & 0.5588 & 0.6346 & 0.652 \\
    \bottomrule
    \end{tabular}}
    \caption{Experimental results of various weight modules on GSM8k benchmark. $W_{up}$ gets the best in three ratios. All weights show the same increasing trend performance.}
    \label{module}
\end{table}

%% file: latex/tza/table/order.tex
\begin{table}[t!]
    \centering
    \resizebox{0.43\textwidth}{!}{
    \begin{tabular}{l|ccc}
    \toprule
    \textbf{Module} & \textbf{25\%}  & \textbf{50\%} & \textbf{75\%} \\
    \midrule
    \text{Original Order} & \textbf{0.6042} & \textbf{0.6497} & \textbf{0.6603} \\
    \cmidrule(lr){1-4}
    \text{Random Order} & 0.5913 & 0.636 & 0.6497 \\
    \text{Min to Max} & 0.6065 & 0.6345 & 0.6436 \\
    \text{Max to Min} & 0.5837 & 0.6322 & 0.6512 \\
    \bottomrule
    \end{tabular}}
    \caption{Inspired by the concept of curriculum learning, we investigated the impact of data order on model performance. We calculated the mean difference scores for the $W_{up}$ weight layer and filtered the data accordingly. The filtered data was then arranged in three ways: from highest to lowest score (Max to Min), from lowest to highest score (Min to Max), and randomly. We compared these arrangements with the original data order.}
    \label{order}
\end{table}

%% file: latex/mxd/algorithm.tex

\begin{algorithm}
\caption{ResoFilter: Parameter Difference-based Data Filtering}
\label{alg:param_diff_filter}
\begin{algorithmic}[1]
\REQUIRE Initial model $M_0$, Training dataset $D$, Filtering ratio $p$
\ENSURE Filtered dataset $D_{filtered}$

\FOR{each sample $d_i \in D$}
    \STATE Fine-tune $M_0$ using $d_i$ to obtain $M_i$
    \STATE $\Delta W \leftarrow M_i - M_0$  \COMMENT{Compute parameter difference}
    \FOR{each layer $l \in M_i$}
        \FOR{each module $m \in l$}
            \STATE $\mathit{diff}_{l,m} \leftarrow \text{mean}(\Delta W_{l,m})$
        \ENDFOR
    \ENDFOR
    
    \STATE $L_{last} \leftarrow \{l_i | i \in [|M_i|-n+1, |M_i|]\}$ \COMMENT{Extract last $n$ layers from $M_i$}

    \STATE $\mathit{diff}_i \leftarrow \text{mean}(\{\mathit{diff}_{l,\text{up\_proj}} | l \in L_{last}\})$
    \STATE Associate $\mathit{diff}_i$ with $d_i$
\ENDFOR

\STATE Sort $D$ in descending order based on $\mathit{diff}_i$
\STATE $k \leftarrow \lfloor |D| \times (1-p) \rfloor$  \COMMENT{Calculate number of samples to retain}
\STATE $D_{filtered} \leftarrow$ Bottom $k$ samples from sorted $D$, preserving original relative order
\COMMENT{This effectively removes the top $p\%$ of samples}

\RETURN $D_{filtered}$
\end{algorithmic}
\end{algorithm}